\documentclass[10pt,twocolumn,letterpaper]{article}

\usepackage{cvpr}
\usepackage{times}
\usepackage{epsfig}
\usepackage{graphicx}
\usepackage{amsmath}
\usepackage{amssymb}

\usepackage{subcaption}
\usepackage{caption}
\usepackage{tabularx}
\usepackage[table]{xcolor}
 \usepackage[normalem]{ulem}
\usepackage{algorithm}
\usepackage[noend]{algpseudocode}
\usepackage{multirow}
\usepackage[section]{placeins}
\usepackage{url}

\usepackage{dblfloatfix}

\usepackage[breaklinks=true,bookmarks=false]{hyperref}

\cvprfinalcopy % *** Uncomment this line for the final submission

\ifcvprfinal\fi

\begin{document}

%%%%%%%%% TITLE
\title{Minimizing Supervision in Multi-label Categorization}

\author{
% Authors at the same institution
Rajat$^{}$\footnotemark[1]\hspace{1.2cm}Munender Varshney$^{}$\footnotemark[1]\hspace{1.2cm}Pravendra Singh\hspace{1.2cm}Vinay P. Namboodiri\\
Department of Computer Science and Engineering, IIT Kanpur, India\\
{\tt\small \{rajatrk, munender, psingh, vinaypn\}@iitk.ac.in}
}

\maketitle
\thispagestyle{empty}

%%%%%%%%% ABSTRACT
\begin{abstract}
   Multiple categories of objects are present in most images. Treating this as a multi-class classification is not justified. We treat this as a multi-label classification problem. In this paper, we further aim to minimize the supervision required for providing supervision in multi-label classification. Specifically, we investigate an effective class of approaches that associate a weak localization with each category either in terms of the bounding box or segmentation mask. Doing so improves the accuracy of multi-label categorization. The approach we adopt is one of active learning, i.e., incrementally selecting a set of samples that need supervision based on the current model, obtaining supervision for these samples, retraining the model with the additional set of supervised samples and proceeding again to select the next set of samples. A crucial concern is the choice of the set of samples. In doing so, we provide a novel insight, and no specific measure succeeds in obtaining a consistently improved selection criterion. We, therefore, provide a selection criterion that consistently improves the overall baseline criterion by choosing the top $k$ set of samples for a varied set of criteria. Using this criterion, we are able to show that we can retain more than 98\% of the fully supervised performance with just 20\% of samples (and more than 96\% using 10\%) of the dataset on PASCAL VOC 2007 and 2012.  Also, our proposed approach consistently outperforms all other baseline metrics for all benchmark datasets and model combinations.
    
\end{abstract}

%%%%%%%%% BODY TEXT
\section{Introduction}
\footnotetext[1]{Equal contribution.}
Deep learning techniques have shown tremendous efficacy in various tasks such as Image classification \cite{Krizhevsky,Simonyan,He}, Object Detection \cite{Girshick1} and image segmentation \cite{Chen,Dai,Long}. We are particularly concerned with the object categorization task that is commonly treated as a multi-class classification task. However, most images commonly contain multiple categories simultaneously being present. Therefore, treating the categorization task as a multi-class classification would not be justified. In fact, it could be treated as such only if we consider an ensemble of binary classifiers, each considering the presence or absence of each individual category. A more natural approach would treat it as a {\it multi-label} classification task. A different way could be using a more structured recognition task such as object detection or segmentation and using full supervision for solving the task. However, a distinct approach is one that considers it as a multi-label categorization task, yet, associates a localization aspect associated with it, such as detection~\cite{Durand_WELDON_CVPR_2016} or segmentation~\cite{Durand_WILDCAT_CVPR_2017}. These methods have shown good performance for classification, indeed, better than most classification methods for obtaining multi-label classification with a smaller number of samples. In this paper, we solve for minimizing the supervision required for realizing these methods.

The localization based multi-label categorization models such as  WELDON and WILDCAT use similar pre-trained backbone (ResNet-101 ~\cite{Simonyan,He} ) to extract features. They use spatial pooling to select relevant regions in the final pooling layer to compute the final score for each class. They show that by localizing positive evidence for an object class, they are able to obtain better performance as compared to standard models that use the entire regions/activation for final prediction. The need for obtaining supervision for all the multiple labels for each instance remains. 
%and the main difference lies in the final pooling layer. These WSL models use what they call a spatial pooling to select relevant regions to compute the final score for each class. This approach allows the WSL models to localize positive evidence for an object class in the given image and provide them with a better performance when compared to standard models that use the entire regions/activation to make their final predictions. The WSL models, though help reduces the burden of bounding box annotated data, still required labeling for all object class present in the image. This is not an easy task and requires significant manual labor, which makes creating these datasets extremely costly, motivating a need to come up with methods to reduce the number of annotations.

The approach proposed by us is to incorporate an active learning strategy to minimize the amount of supervision required. In active learning, based on the current model, one chooses a set of samples that need to be labeled. The ground-truth labels for these samples are obtained by an annotator, and the model is further retrained to obtain the next of samples. The selection of samples thus forms the core element in any active learning strategy. Unfortunately, our experiments suggest that no particular strategy or measure for active learning consistently suffices to provide a set of samples. To address this drawback, we provide a simple strategy that is performant. We choose a varied set of active learning methods, such as those based on entropy and classification uncertainty. We augment these strategies with a couple of strategies that consider the localization aspects obtained through the specific localization based categorization methods. Given this set of strategies, if we need to choose $n$ samples given $k$ strategies, we obtain the ordered set of top $n/k$ samples that are unique using the $k$
strategies. We show that this simple method outperforms any individual strategy, costs similar to any single baseline strategy, and is also better than other variants for obtaining an ensemble. We term this strategy a `metric agnostic' sampling strategy. 
%In this paper, we have proposed two active learning approaches to further reduce this annotation cost in the Weakly Supervised Learning (WSL) models for multi-label object recognition. The first novel active learning approach is based on measuring the capability of the WSL model to separate the foreground from the background for a data sample, which is termed as the foreground-background separation metric. The other proposed ``metric agnostic" approach is used to provide the subset of data based on the novel method to ensemble the existing active learning approaches to get a performance that is better than each of the approaches used. The ``metric agnostic" approach performs consistently well over all datasets and models combination irrespective of the other active learning metric, which might not perform in some cases during evaluation. Experimentation on benchmark dataset shows using the foreground-background separation metric helps models perform better compared to other benchmark metrics, especially with a relatively small amount of data. Also, evaluation of the ``metric agnostic" approach on different models and benchmark datasets, shows it outperforming all other metrics in any setting. We also conduct ablation analysis, which demonstrates that the proposed novel technique to ensemble various active learning approaches is not trivial, and it outperforms other ways of ensemble methods.   

% not only superior to variants but is also computationally comparable to other methods.

\section{Related Work}
Object recognition and detection have been extensively studied and achieved state-of-the-art results for the deep learning models such as CNN ~\cite{Durand_WILDCAT_CVPR_2017,Girshick,Girshick1,liu2016ssd,Durand_WELDON_CVPR_2016,Krizhevsky}. The earlier method in object detection using deep learning is based on explicitly generating region proposals, following by feature extraction, and classification \cite{Girshick1,ren2015faster}. The newer method outputs the location of the object and class directly without any intermediate proposal stage, such as SSD \cite{liu2016ssd} and YOLO \cite{redmon2016you}. Due to these marvelous performances, CNN models have been extensively used to solve multi-label object recognition problem by localization using region proposal \cite{liu2016ssd,Girshick1,Girshick}. However, localization in CNN models needs bounding box annotation for large dataset \cite{Girshick,liu2016ssd,Girshick1}, which is a laborious and costly job. Due to this laborious work, the weakly supervised learning models got huge attention \cite{Durand_WILDCAT_CVPR_2017,Durand_WELDON_CVPR_2016}. These models use spatial pooling with Fully convolutional networks, which helps in localizing discriminative regions \cite{zhang2018adversarial,Durand_WILDCAT_CVPR_2017}. But, the annotation cost to label all objects in the dataset is huge, so active learning is used to label a smaller set of images.   

Active learning has been wide used in various computer vision task where the labeled data is scarce and difficult to obtain such as image classification, segmentation, and  detection \cite{islam2016active}, automated and manual video annotation \cite{karasev2014active}, and CAPTCHA recognition \cite{stark2015captcha}, segmentation\cite{islam2016active,kapoor2010gaussian}.
Specifically active learning has also helped in object classification \cite{islam2016active,kapoor2010gaussian,yang2015multi,joshi2009multi} and object detection \cite{brust2018active}. These tasks have been explored for various specific applications where annotated data is scarce, such as vehicle images \cite{sivaraman2014active} and satellite images \cite{bietti2012active}. A detailed survey of active learning algorithms, which are used to select fewer labeled training data for achieving high accuracy, are present in the settles's survey \cite{settles.tr09}. The approaches developed so far doesn't include any active learning work (multi-label classification) for deep learning settings. Active learning concepts relevant for this work include uncertainty sampling \cite{stark2015captcha,islam2016active,yang2015multi} and query-by-committee \cite{seung1992query}. These active learning methods have helped in solving computer vision problems. These method also find application in different learning models including Gaussian processes \cite{kapoor2010gaussian}, decision trees \cite{lewis1994heterogeneous} and SVMs \cite{tong2001support}. 

\begin{figure}
    \centering
    \includegraphics[width=0.9\linewidth]{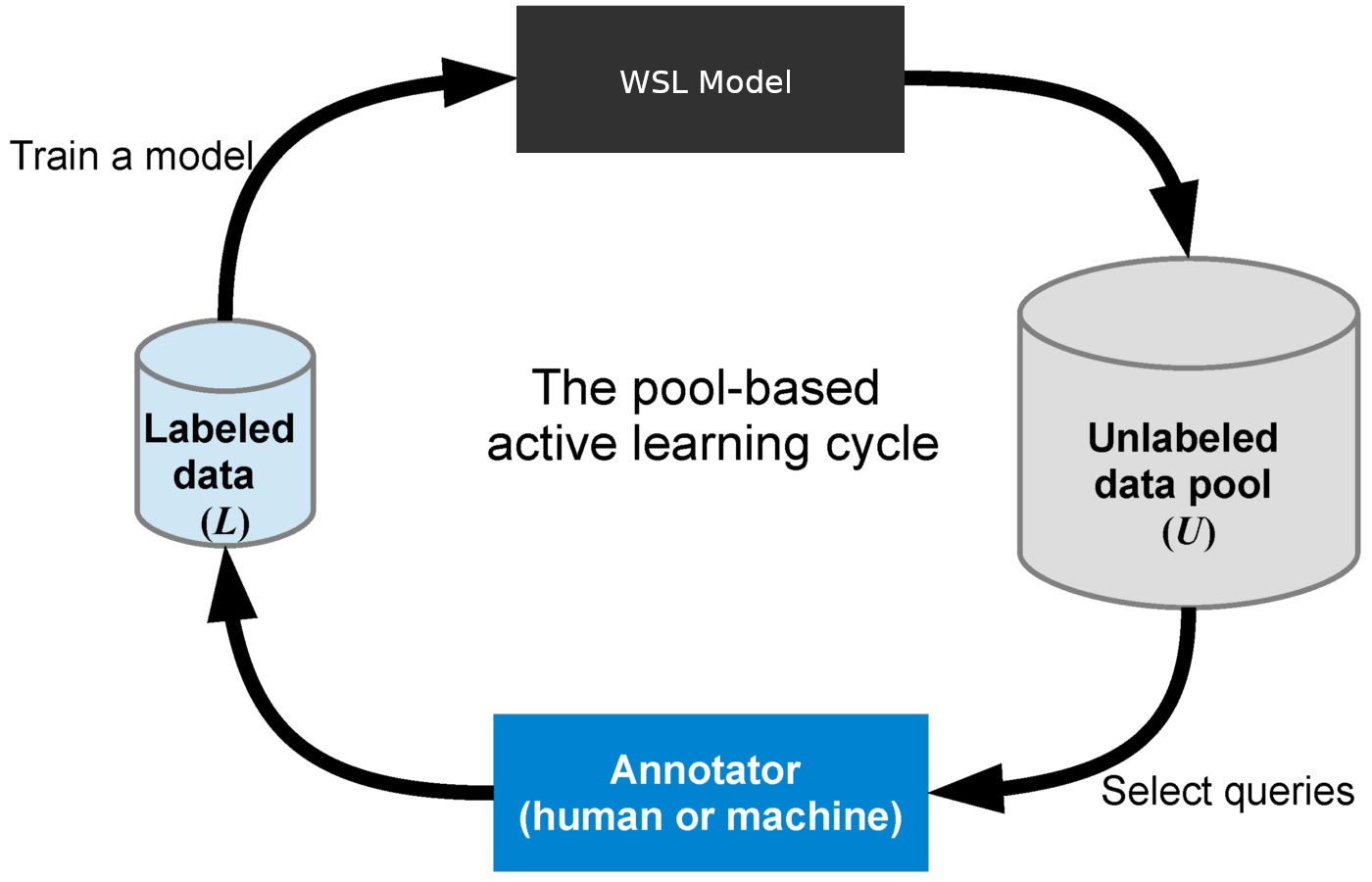}
    \caption{An iteration of active learning.}
    % \vspace{-15pt}
    \label{figure1}
\end{figure}

Even though active learning has been used in machine learning for decades, it has only recently been applied to deep learning architectures.  Wang \emph{et. al.,} in \cite{wang2017cost}  and \cite{wang2014new}  has proposed to use uncertainty based active learning for deep network model. Several metrics are offered to estimate the uncertainty in the model, such as margin, least confidence, or entropy sampling. Starket \emph{et. al.,} \cite{stark2015captcha} proposed another variation of margin sampling. Another work in paper \cite{liu2017active} presents uncertainty sampling. In addition, a query by committee strategy, as well as weighted incremental dictionary learning for active learning, is proposed. Gal \emph{et al.} \cite{gal2017deep} used Monte Carlo dropout to measure Bayesian uncertainty for selecting samples in CNN models. To the best of our knowledge, this is the first work that addresses the issue of multi-label object recognition in a deep learning setting using active learning. However, there are some good work has been done non-deep setting for multi-label active learning which leverage than the correlation between labels \cite{li2013active,huang2015multi,du2017robust}.%------------------------------------------------------------------------
\vspace{10pt}
\section{Active Learning for WSL (Weakly Supervised Learning) models}

A trained multi-label classifier takes an input image and outputs the probability of the presence of one or more objects of a class for each of the predefined categories. To train this model, we need images that are annotated on the global level, i.e., class-level annotations. Such annotations are available in public datasets like PASCAL VOC \cite{Everingham10} and MS COCO \cite{COCO}. We implement an active learning algorithm in the way shown in Fig. \ref{figure1}, where the baseline model is trained with a small amount of annotated samples. The baseline model is then used over the unlabelled dataset to give a pool of most informative samples using the given active learning metric. These pooled unlabelled samples are then annotated using human annotators and then added to the training set to train a new model. This cycle is an iteration of an active learning algorithm, which can be repeated multiple times until the annotation capability or the desired level of performance is reached.
\par

We first take a look at our proposed foreground-background active learning metric in Sec. \ref{section3.1}, which utilizes the WSP module of the WELDON \cite{Durand_WELDON_CVPR_2016} and the multi-map WSL transfer layer of WILDCAT \cite{Durand_WILDCAT_CVPR_2017} model to select the most informative samples from the unlabelled data pool. After that, we look at the metric agnostic approach (Sec. \ref{section3.2}), which utilizes existing active learning metrics to pool the best samples.

\subsection{Foreground Background Separation} \label{section3.1}
The main idea for this active learning metric utilizes the way in which these WSL models, both WELDON \cite{Durand_WELDON_CVPR_2016} and WILDCAT \cite{Durand_WILDCAT_CVPR_2017} are trained and how they perform classification.

% \begin{figure*}
% \begin{center}
% \fbox{\rule{0pt}{2in} \rule{.9\linewidth}{0pt}}
% \end{center}
%   \caption{Figure for WELDON architecture.}
% \label{figure2}
% \end{figure*}

\par

In the WELDON architecture, the network consists of two parts, and the first part acts as a feature extractor having a CNN like ResNet \cite{He} or VGG \cite{Simonyan} trained on a large scale dataset such as ImageNet \cite{ImageNet}, and the second part which contains the WSP module which uses final convolution layer followed by a spatial pooling to make classification predictions. This final WELDON spatial pooling, which improves upon the MIL \cite{MIL} and gets insights from it, pools both top instances as well as bottom instances, after which it averages these values to obtain the final score. The reason to perform such operation, being that the top instances correspond to localized positive evidence for the presence of an object, and the bottom instances correspond to negative evidence for its absence. Thus, the model's ability to make a correct prediction for the presence/absence of an object is dependent on its ability to discern the foreground (localized positive evidence) from the background (negative evidence) in the given sample. Thus, we reason that samples in which the model can easily discern the foreground from the background, which here is being quantified by the difference between the value of maximum top instances and minimum negative instances, will not contain much new information for the model to learn from. We use this reasoning to propose an active learning metric which selects samples based on how well the model can discern foreground from the background instead of using the model's final output like the other traditional active learning metrics. 

% \begin{figure*}
% \begin{center}
% \fbox{\rule{0pt}{2in} \rule{.9\linewidth}{0pt}}
% \end{center}
%   \caption{Figure for WILDCAT architecture.}
% \label{figure3}
% \end{figure*}

\par
The proposed Foreground-Background approach can be applied over WSL models or similar kinds of architecture only, i.e., WILDCAT architecture. The WILDCAT architecture, which similar to the WELDON architecture, also has two parts a feature extractor and a prediction network. The latter contains the multi-map WSL transfer layer. Unlike the WELDON network, WILDCAT uses two different kinds of pooling to make the final prediction. These pooling methods include a class-wise pooling, which is then followed by a spatial pooling, latter of which is similar to that used in WELDON. In this network architecture, we again quantify the foreground and background separation as defined above, using the tensor/output obtained after performing the class-wise pooling on the WSL transfer layer. The proposed Foreground-Background approach is only limited by choice of architecture and not the task. 

Using the metric as explained above, we select samples to annotate from the unlabelled dataset for which the model performance on the foreground and background separation is weak, i.e., the samples for which difference between the maximum of top and minimum of bottom instances are small. We also use different aggregation techniques for combining this foreground-background separation performance across different classes. For more detail, see Sec. \ref{section5}.
%------------------------------------------------------------------------
\begin{algorithm}
\caption{Metric Agnostic Approach}\label{alg:metricagnostic}
\begin{algorithmic}[1]
\State \textbf{Input:} \\ $S = [S_{M1}, S_{M2}, S_{M3}, \dots, S_{MT}]$ where $S_{Mi}$ is list of selected samples sorted using metric $Mi $ \\ Number of samples to be selected $n$
\State \textbf{Output}\\ List of samples given by metric agnostic approach,$ \mathcal{S}$.
\State\textbf{Algorithm:}
\State Create an empty set, $\mathcal{S} = \phi$
\State Create an index list, idxs $ = [1, 1, \dots, 1]$ 
\While{$|\mathcal{S}| < n$}
\For{ $t = 1$ to $T$}
\If{idxs$[t] \leq len( S_{Mt} )$}
\State $\mathcal{S} = \mathcal{S} \cup S_{Mt}$[idxs[t]]
\State idxs$[t]++$
\EndIf
\If{$|\mathcal{S}| \geq n$}
\State \textbf{break}
\EndIf
\EndFor
\EndWhile
\State \textbf{return} $\mathcal{S}$
\end{algorithmic}
\end{algorithm}

\subsection{Metric Agnostic Approach (AG)} \label{section3.2}
%In the course of our experimentation for evaluating the performance of our foreground-background separation metric and comparing it to other benchmark active learning metrics (see Sec. \ref{section4}), we observed that the best performing active learning metric varied with a change in the model, the dataset and even the amount of data being used to train the model. Different active learning metric seems to outperform each other depending upon the combination of model, dataset, and amount of data being used.
From the evaluation of various active learning techniques, we observed that the performance of these techniques (metrics) depends on various conditions (amount of data, dataset, and model). We, therefore, propose a new method that could combine the best set of samples obtained from each metric. This new method we term as being a ``metric agnostic" approach as it can be thought of as an algorithm that is based on other active learning metrics. 

\par

%This inspired us to come up with a ``best" active learning metric, which performs at least as good as the best performing active learning metric for any dataset, model, and amount of training data combination. We came up with what we call a metric agnostic approach, where we used the foreground-background separation metric (Sec. \ref{section3.1}) along with other benchmark active learning metrics to pool samples from the unlabeled dataset. In this metric agnostic approach, 
In the proposed metric agnostic approach, the main idea of the proposed method is to pool samples from samples selected using other benchmark active learning metrics. Specifically, we take the top picks of each of the individual active learning metrics in equal proportion and then annotate these aggregated samples and then add them to the training dataset. The way we pool samples from different metrics is we start with lists of samples selected by each of the metrics, sorted on some priority based on how that metric selects the samples. We then start with an empty set, and for each iteration, we select one sample from each of the list, which has not already been selected (by some other metric) until we have reached our predefined limit as shown in Algorithm \ref{alg:metricagnostic}. The final set of unlabelled samples obtained this way is considered as the final selection of the metric agnostic approach (see Alg. \ref{alg:metricagnostic}).

It should be noted that though this paper shows the usage of the metric agnostic approach with WSL models for multi-label classification. But, the metric agnostic approach is neither bound to the model type nor the task at hand. Hence it can be used with all kinds of models and tasks. Exploration of this capability of the metric agnostic approach will be done as part of future work.

%------------------------------------------------------------------------

%------------------------------------------------------------------------
\section{Baseline Active Learning Metrics}\label{section4}
We compare our proposed active learning metrics against various other widely used benchmark active learning metrics for the classification task, as well as with passive learning. We define $P(y_{j}\mid x_{i})$ as the model's confidence for the presence of class $j$ in the $i^{th}$ sample, these confidence values are obtained by taking the sigmoid of the model's output.

\subsection{Classification Uncertainty (UNC)}
As defined $P(y_{j}\mid x_{i})$ denotes the model confidence for the presence of class j in the $i^{th}$ sample, we can measure the model's certainty for class $j$ as $abs(P(y_{j}\mid x_{i}) - 0.5)$, as a value of $P(y_{j}\mid x_{i})$ close to $0$ denotes that the model is certain the class is not present and a value of $P(y_{j}\mid x_{i})$ close to $1$ denotes that model is certain that the class is present. We then aggregate these certainties value across different classes using sum operator and select samples with following strategy $\min_i\sum_jabs(P(y_{j}\mid x_{i}) - 0.5)$.
%------------------------------------------------------------------------

\subsection{Maximum Entropy (ENT)}
In this active learning metric we select samples with maximum entropy values. The query strategy is defined as: $\max_i-\sum_jP(y_{j}\mid x_{i})\log{P(y_{j}\mid x_{i})}$
%------------------------------------------------------------------------

\subsection{Min-Max (MM)}
In this active learning metric, we select samples with minimum-maximum model confidence across all classes, defined as $\min_i\max_jP(y_j\mid x_i)$ 
%------------------------------------------------------------------------

\subsection{Passive Learning (R)}
In this strategy, we randomly select samples from an unlabeled dataset, annotate them, and add to the training set.
%------------------------------------------------------------------------

%------------------------------------------------------------------------

\section{Experiments} \label{section5}
We evaluate our proposed metrics on both WELDON \cite{Durand_WELDON_CVPR_2016} and WILDCAT \cite{Durand_WILDCAT_CVPR_2017} models. All our experiments were performed using the PyTorch framework. For all our experiments we used authors implementation of the model which are available online, WELDON (https://github.com/durandtibo/weldon.resnet.pytorch/) and WILDCAT (https://github.com/durandtibo/wildcat.pytorch/). The final results for each of the metrics are obtained by performing three trials using three different sets of randomly selected images to train the baseline model. Also, for the baseline and the first three of the five iterations, the model was trained for $35$ epochs, and for the last two iterations, the model was trained for $25$ epochs. For each active learning metric, we show relative(\%) mAP of the model to that trained on the entire dataset in each graph. 

\par

For the foreground-background separation metric described in Sec. \ref{section3.1}, we use three different aggregation methods for experimentation.
\begin{itemize}
    \item \textbf{SEPSUM}, the performance of the model on a sample is represented using the sum of the foreground-background separation value of all classes.
    \item \textbf{SEPMAX}, the performance of the model on a sample is represented by the maximum value of foreground-background separation across all classes.
    \item \textbf{SEPMIN}, the performance of the model on a sample is represented by the minimum value of foreground-background separation across all classes. 
\end{itemize}

For each of the variants, we select those samples from the unlabelled dataset on which the model's performance is poor, i.e., the samples with a minimum value of \textbf{SEPSUM}, \textbf{SEPMAX} and \textbf{SEPMIN}. Finally, all the experiments with the metric agnostic approach are performed, as explained in Sec. \ref{section3.2}.

\par

\textbf{Datasets: } The performance of our proposed active learning approach has been evaluated on three benchmark datasets such as  PASCAL VOC 2007, PASCAL VOC 2012, and MS COCO \cite{Everingham10, COCO}. For all classification experiments protocols used in \cite{Durand_WELDON_CVPR_2016} were followed. For each of these datasets, we start with a baseline model trained on a small number of randomly selected images from the training set, and the remaining training dataset is then used as the pool of unlabelled images from which we pool images in small increment for multiple iterations using active learning metrics. Since the annotations for the training set of each of the datasets are available, we provide the annotations for the selected images without using human annotators. For each of the experiments, we use up to $20\%$ of the training dataset to train the model and then compare its performance with the model trained on the entire training set.

\subsection{PASCAL VOC 2007}\label{section5.1}
\textbf{Experimental Setup:} For Pascal VOC 2007 dataset, we begin with a baseline model trained on $100$ randomly selected images from the training set (containing $\sim 5000$ images). Then for each active learning metric (excluding passive learning case), we add new images in five iterations adding respectively $100$, $100$, $100$, $100$, and $500$ images in each iteration training the final model using $1000$ images.

\begin{figure}
    \centering
    \includegraphics[width=0.8\linewidth]{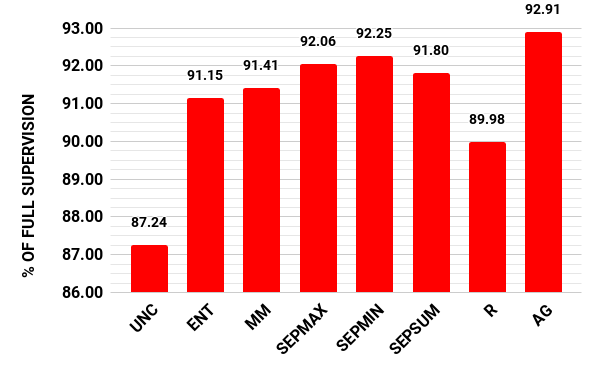}
    % \vspace{-5pt}
    \caption{Results for PASCAL VOC 2007 using $500$ images and WELDON. For each active learning metric we show relative(\%) mAP of model to that trained on entire dataset.}
    % \vspace{-10pt}
    \label{figure4}
\end{figure}

\begin{figure}
    \centering
    \includegraphics[width=0.8\linewidth]{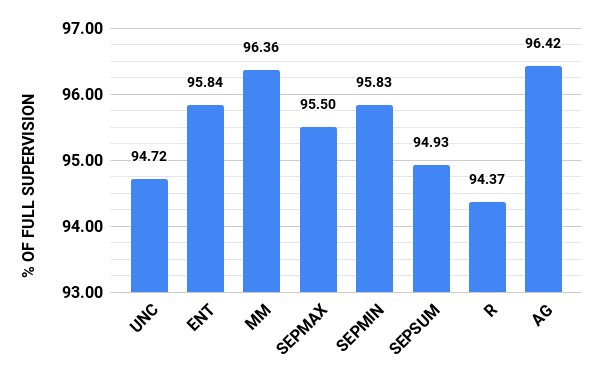}
    %  \vspace{-5pt}
    \caption{Results for PASCAL VOC 2007 using $1000$ images and WELDON.}
    % \vspace{-10pt}
    \label{figure5}
\end{figure}
\begin{figure}
    \centering
    \includegraphics[width=0.8\linewidth]{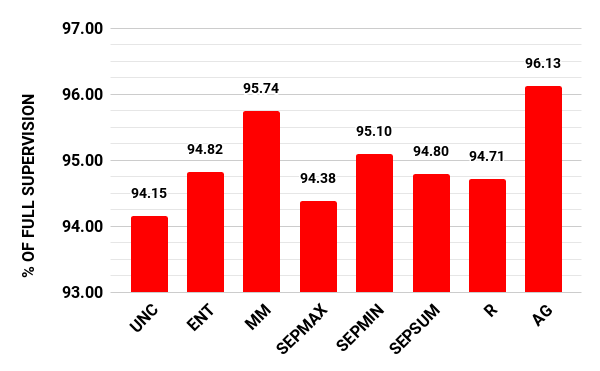}
    % \vspace{-5pt}
    \caption{Results for PASCAL VOC 2007 using $500$ images and WILDCAT.}
    %  \vspace{-10pt}
    \label{figure6}
\end{figure}
\begin{figure}
    \centering
    \includegraphics[width=0.8\linewidth]{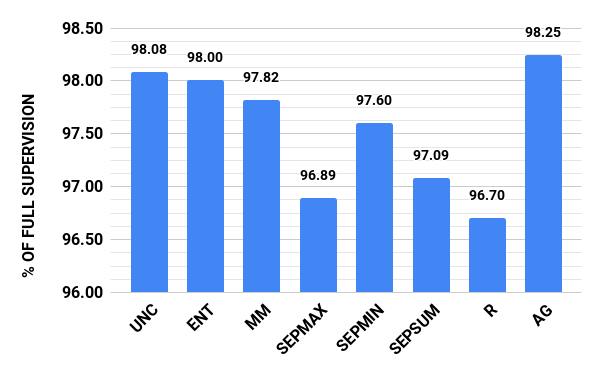}
    %  \vspace{-5pt}
    \caption{Results for PASCAL VOC 2007 using $1000$ images and WILDCAT.}
    %  \vspace{-10pt}
    \label{figure7}
\end{figure}

\par

\textbf{Results:} For WELDON using $500$ images in Fig. \ref{figure4} and using $1000$ images in Fig. \ref{figure5}. For WILDCAT using $500$ images in Fig. \ref{figure6} and using $1000$ images in Fig. \ref{figure7}. The images show the performance of the model using each of the active learning metrics relative to the performance of the model trained on the entire training dataset. 

\par
We have three main observations from the results obtained on the PASCAL VOC 2007 dataset, first being that when using $500$ images, our proposed foreground-background separation metric (in this case \textbf{SEPMIN}) outperforms all other metrics (except for metric agnostic) while using WELDON network. This metric helps the WSL model to perform much better while using a relatively small amount of data. Secondly, the performance of active learning metrics varies greatly with change in model, as evident with the performance of uncertainty(\textbf{UNC}) metric which goes from being the worst-performing metric while using WELDON (Fig. \ref{figure5}) to the best performing metric (except for metric agnostic) while using the WILDCAT model (Fig. \ref{figure7}). We also observe a significant difference in performance as a result of the change in the amount of data while using the same model. For the case of both WELDON (Fig \ref{figure4} and Fig. \ref{figure5}) and WILDCAT (Fig. \ref{figure6} and Fig. \ref{figure7}) comparative performance of all the metrics changes significantly as we go from $500$ images to $1000$ images. The third important observation is the consistent performance of our metric agnostic approach across all cases. The metric agnostic approach outperforms all other active learning metrics, especially for the case when 500 images are being used to train the model. This goes to show that apart from the consistent performance across different scenarios, the metric agnostic approach, when used in scenarios with limited annotation capability, can help get a non-trivial improvement over the performance obtained from using any single active learning metric.
%------------------------------------------------------------------------

\subsection{PASCAL VOC 2012}\label{section5.2}
\textbf{Experimental Setup:} For VOC 2012 dataset we begin with a baseline model trained on $95$ randomly selected images from the training set (containing $\sim 5700$ images). Then for each active learning metric (excluding passive learning case) we add new images in six iterations adding respectively $95$, $95$, $95$, $95$, $95$ and $570$ images, training the final model using $1140$ images. Final results is reported similar to VOC 2007 Sec. \ref{section5.1}

\begin{figure}
    \centering
    \includegraphics[width=0.8\linewidth]{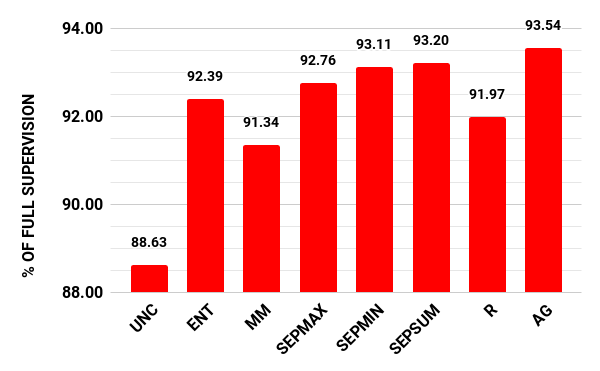}
    \caption{Results for PASCAL VOC 2012 using $570$ images and WELDON.}
    %  \vspace{-8pt}
    \label{figure8}
\end{figure}

\begin{figure}
    \centering
    \includegraphics[width=0.8\linewidth]{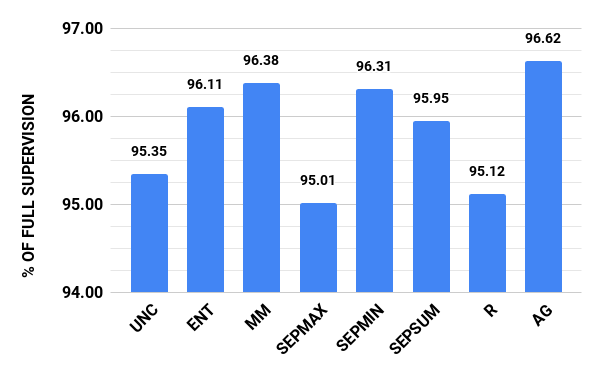}
    \caption{Results for PASCAL VOC 2012 using $1140$ images and WELDON.}
    %  \vspace{-8pt}
    \label{figure9}
\end{figure}
\begin{figure}
    \centering
    \includegraphics[width=0.8\linewidth]{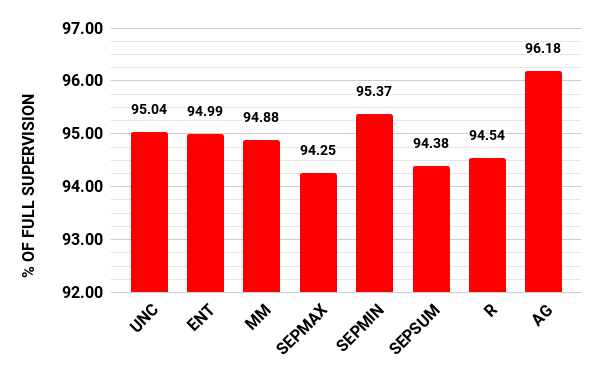}
    \caption{Results for PASCAL VOC 2012 using $570$ images and WILDCAT.}
    %  \vspace{-12pt}
    \label{figure10}
\end{figure}
\begin{figure}
    \centering
    \includegraphics[width=0.8\linewidth]{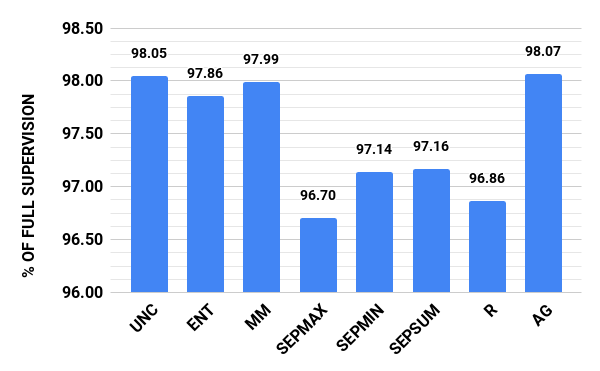}
    \caption{Results for PASCAL VOC 2012 using $1140$ images and WILDCAT.}
    %  \vspace{-12pt}
    \label{figure11}
\end{figure}

\par

\textbf{Results:} For WELDON using $570$ images in Fig. \ref{figure8} and using $1140$ images in Fig. \ref{figure9}. For WILDCAT using $570$ images in Fig. \ref{figure10} and using $1140$ images in Fig. \ref{figure11}. Similar to the case of the PASCAL VOC 2007 case, when using $570$ images, our proposed foreground-background separation metric outperforms all other metrics (except for metric agnostic) not only while using WELDON network (\textbf{SEPSUM}) but also for the case for WILDCAT network (\textbf{SEPMIN}). This further makes a case for the suitability of this metric is for WSL networks. The performance of active learning metrics again varies with the change in model and the amount of data being used to train the model. Also similar to VOC 2007 case, the metric agnostic approach consistently outperforms all other active learning metrics, especially for the case when using $570$ images to train the model. 
%------------------------------------------------------------------------

\subsection{MS COCO}
\textbf{Experimental Setup:} For the MS COCO dataset, we begin with a baseline model trained on $1640$ randomly selected images from the training set (containing $\sim 82000$ images). Then for each of the active learning metrics (excluding passive learning case), we add new images in five iterations adding respectively $1640$, $1640$, $1640$, $1640$, and $8200$ images, training the final model using $16400$ images.

\begin{figure}
    \centering
    \includegraphics[width=0.8\linewidth]{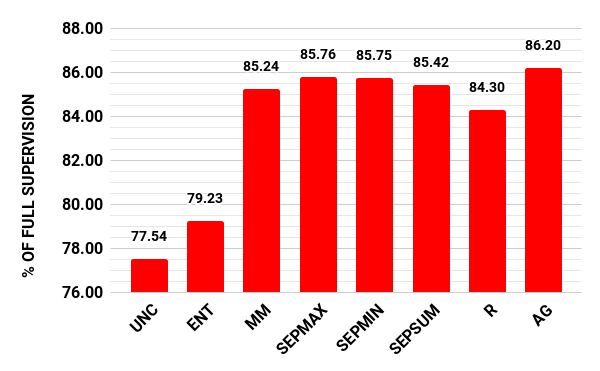}
    \caption{Results for MS COCO using $8200$ images and WELDON.}
    %  \vspace{-12pt}
    \label{figure12}
\end{figure}

\begin{figure}
    \centering
    \includegraphics[width=0.8\linewidth]{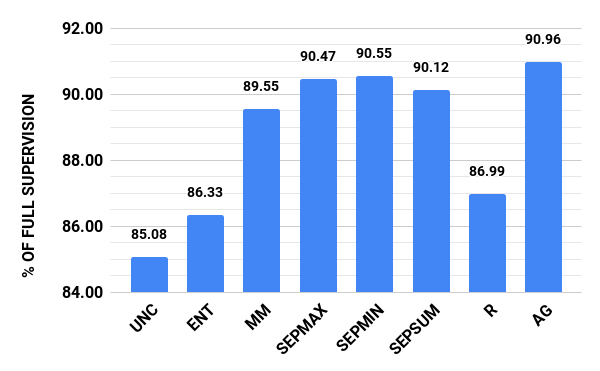}
    \caption{Results for MS COCO using $16400$ images and WELDON.}
    %  \vspace{-12pt}
    \label{figure13}
\end{figure}
\begin{figure}
    \centering
    \includegraphics[width=0.8\linewidth]{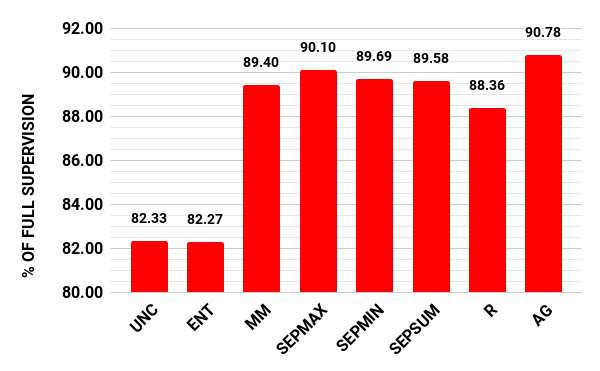}
    \caption{Results for MS COCO using $8200$ images and WILDCAT.}
    %  \vspace{-12pt}
    \label{figure14}
\end{figure}
\begin{figure}
    \centering
    \includegraphics[width=0.8\linewidth]{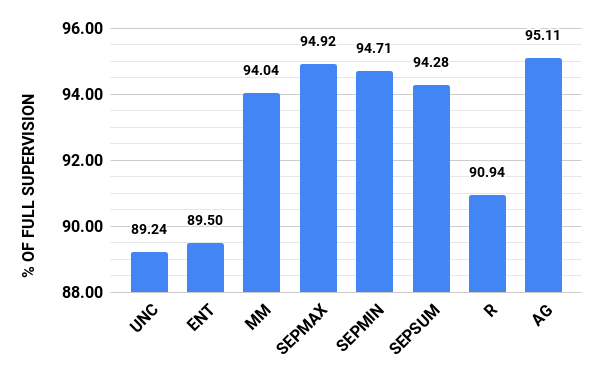}
    \caption{Results for MS COCO using $16400$ images and WILDCAT.}
    %  \vspace{-12pt}
    \label{figure15}
\end{figure}
\par

\textbf{Results:} For WELDON using $8200$ images in Fig. \ref{figure12} and using $16400$ images in Fig. \ref{figure13}. For WILDCAT using $8200$ images in Fig. \ref{figure14} and using $16400$ images in Fig. \ref{figure15}. When using $8200$ images, our proposed foreground-background separation metric (\textbf{SEPMAX}) outperforms all other metrics (except for metric agnostic) not only while using WELDON network but also for the case for WILDCAT network. Also, when using $16400$ images our proposed foreground-background separation metric outperforms all other metrics (except for metric agnostic) for both WELDON (\textbf{SEPMIN}) and WILDCAT (\textbf{SEPMAX}). This goes to show that our proposed foreground-background metric is able to scale well for large datasets. Similar to what we observed for VOC datasets, the performance of active learning metrics again varies with the change in model and amount of data being used to train the model. Also, similar to the VOC datasets case, the metric agnostic approach consistently outperforms all other active learning metrics. 
%------------------------------------------------------------------------

\subsection{Ablation Study}
\begin{figure}
    \centering
    \includegraphics[width=0.8\linewidth]{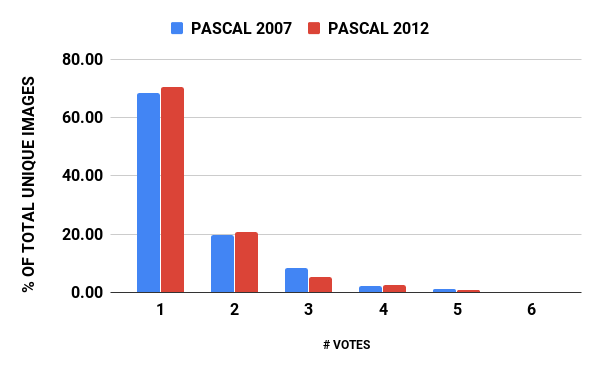}
    \caption{Vote distribution obtained using WELDON for both PASCAL VOC 2007 and 2012 dataset. Active learning metrics being used are \textbf{UNC}, \textbf{ENT}, \textbf{MM}, \textbf{SEPMAX}, \textbf{SEPMIN} and \textbf{SEPSUM}.}
    %  \vspace{-12pt}
    \label{figure16}
\end{figure}

\begin{figure}
    \centering
    \includegraphics[width=0.8\linewidth]{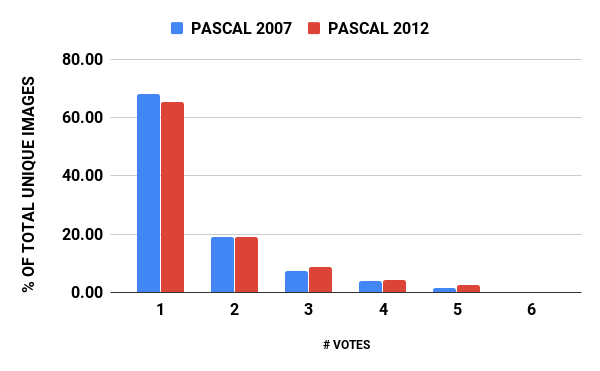}
    \caption{Vote distribution obtained using WILDCAT for both PASCAL VOC 2007 and 2012 dataset. Active learning metrics being used are \textbf{UNC}, \textbf{ENT}, \textbf{MM}, \textbf{SEPMAX}, \textbf{SEPMIN} and \textbf{SEPSUM}.}
    %  \vspace{-12pt}
    \label{figure17}
\end{figure}

Various ablation studies have been conducted to illustrate various aspects of the proposed metric agnostic approach. First, We compare our aggregation strategy of the metric agnostic method explained in Sec. \ref{section3.2}, to another voting based aggregation approach in which for each of the sample selected by one or more active learning metric, we treat the number of active learning metrics that selected that sample as votes for that sample, i.e., sample selected by only one metric will get one vote and so on. Finally, samples are selected based on the decreasing order of the number of votes. The vote distribution we obtained using WELDON is shown in Fig. \ref{figure16} and using WILDCAT is shown in Fig. \ref{figure17}. From the results obtained, we see that more than half of the samples receive just one vote suggesting that, for the most part, there is little agreement between different active learning metrics on the importance of a sample. Furthermore, the classification performance obtained using this voting approach was sub-par to that obtained using our proposed approach. 

\par


The second ablation study is to show that though our metric agnostic approach is an ensemble-based, the additional time cost for this ensemble is relatively quite low. The computation for each of the active learning metrics being used under it can be done with a single pass through the network. Table \ref{table1} shows the results of our time analysis using WELDON using an NVIDIA GeForce GTX 1080 graphics card. Table \ref{table2} shows result for WILDCAT model. For both models, we are pooling $100$ images from the PASCAL VOC 2007 dataset(from $\sim 4900$ images), and for PASCAL VOC 2012, we are pooling $95$ images (from $\sim5600$ images).

\begin{table}
\begin{center}
\renewcommand{\arraystretch}{1.0}
\begin{tabular}{|c|c|c|}
\hline
Dataset & VOC 2007 & VOC 2012 \\
\hline\hline
\textbf{UNC} (sec) & 40.888 & 50.415\\
\textbf{ENT} (sec) & 40.868 & 50.335\\
\textbf{MM} (sec) & 40.771 & 50.429\\
\textbf{SEPMAX} (sec) & 41.557 & 51.328\\
\textbf{SEPMIN} (sec) & 41.728 & 51.442\\
\textbf{SEPSUM} (sec) & 41.490 & 51.496\\
\textbf{AG} (sec) & 43.385 & 53.454\\
$\Delta_{\textbf{AG}} (\%)$ & 5.259 & 5.002\\
\hline
\end{tabular}
\end{center}
\caption{Time analysis of active learning metrics for both PASCAL VOC 2007 and 2012  using WELDON. $\Delta_{\textbf{AG}}$ is the relative(\%) time difference between \textbf{AG} and average of all other metrics.}
% \vspace{-8pt}
\label{table1}
\end{table}

\begin{table}
\begin{center}
\renewcommand{\arraystretch}{1.0}
\begin{tabular}{|c|c|c|}
\hline
Dataset & VOC 2007 & VOC 2012 \\
\hline\hline
\textbf{UNC} (sec) & 101.966 & 120.388\\
\textbf{ENT} (sec) & 101.756 & 120.467\\
\textbf{MM} (sec) & 102.035 & 120.338\\
\textbf{SEPMAX} (sec) & 101.916 & 120.971\\
\textbf{SEPMIN} (sec) & 102.220 & 120.957\\
\textbf{SEPSUM} (sec) & 102.755 & 121.367\\
\textbf{AG} (sec) & 106.557 & 124.166\\
$\Delta_{\textbf{AG}} (\%)$ & 4.357 & 2.831\\
\hline
\end{tabular}
\end{center}
\caption{Time analysis of active learning metrics for both PASCAL VOC 2007 and 2012 using WILDCAT. $\Delta_{\textbf{AG}}$ is the relative(\%) time difference between \textbf{AG} and average of all other metrics.}
% \vspace{-8pt}
\label{table2}
\end{table}

\begin{figure}
    \centering
    \includegraphics[width=0.8\linewidth]{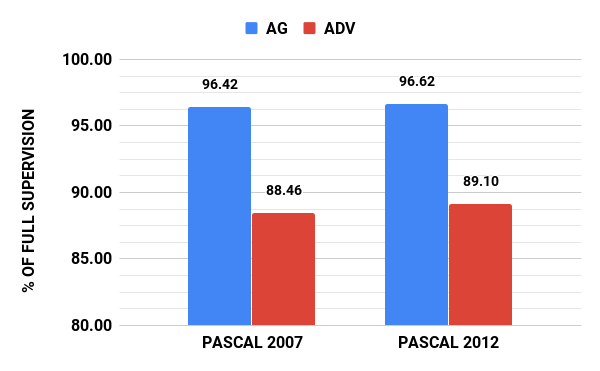}
    \caption{Performance comparison of Metric Agnostic with Adversarial on both PASCAL VOC 2007 (using $1000$ images) and 2012 (using $1140$ images) using WELDON.}
    % \vspace{-8pt}
    \label{figure18}
\end{figure}

\begin{figure}
    \centering
    \includegraphics[width=0.8\linewidth]{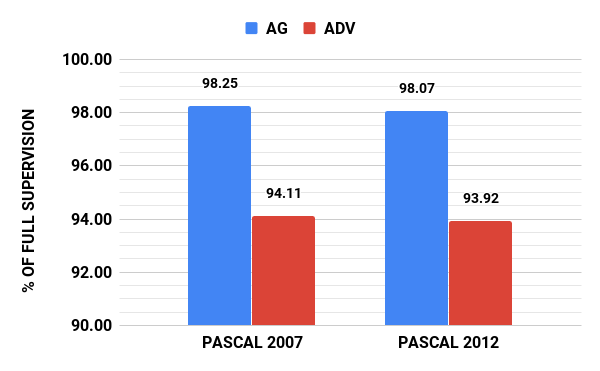}
    \caption{Performance comparison of Metric Agnostic with Adversarial on both PASCAL VOC 2007 (using $1000$ images) and 2012 (using $1140$ images) using WILDCAT. For both method we show relative(\%) mAP of model to that trained on entire dataset.}
    % \vspace{-8pt}
    \label{figure19}
\end{figure}

The third ablation study is to show that effective selection of samples based on model behavior is critical and cannot be replaced by ``naively" selecting samples that contain the most number of classes to maximize the number of examples of each class a model sees. To test this hypothesis, we created a sort of  ``adversarial"(\textbf{ADV}) dataset in which samples are added in decreasing order of the number of classes it contains. The performance comparison of this adversarial method with our metric agnostic method is shown in Fig. \ref{figure18} and Fig. \ref{figure19}, for both cases, we see that the metric agnostic method outperforms the adversarial method. 

%------------------------------------------------------------------------
\section{Conclusions}
The two active learning metrics are presented for multi-label object recognition in WSL (Weakly Supervised Learning) models. Our foreground-background separation metric pools images based on the model's performance on being able to discern foreground from the background. Our second metric agnostic metric utilizes the existing active learning metrics to make its selection. Through extensive experimentation, we then show that the foreground-background separation metric outperforms existing active learning metrics, especially when the amount of annotations available is limited and that this metric scales well for the case of large datasets like MS COCO. The metric agnostic approach is shown to consistently outperform all other active learning metrics for various models and dataset combinations, indicating its robustness to change in datasets and models. Future works include adapting these proposed metrics to other domains of visual tasks like semantic segmentation.

{\small
\bibliographystyle{ieee_fullname}
\bibliography{camera_ready}
}

\end{document}